\documentclass[submission,copyright,creativecommons]{eptcs}
\providecommand{\event}{Thedu'17} % Name of the event you are submitting to

\usepackage{underscore}
\usepackage{graphicx}
\usepackage{hyperref}
\usepackage{url}
\usepackage{soul}
\usepackage{color}
\usepackage[dvipsnames]{xcolor}
\usepackage[english]{babel}
\usepackage{algorithmic}

\usepackage[T1]{fontenc}
\usepackage[utf8]{inputenc}

\title{Improving QED-Tutrix by Automating the Generation of Proofs}
\author{Ludovic Font
\institute{\'Ecole Polytechnique de Montr\'eal\\ Montr\'eal, Qc, Canada}
\email{ludovic.font@polymtl.ca}
\and
Philippe R. Richard
\institute{Universit\'e de Montr\'eal\\
Montr\'eal, Qc, Canada}
\email{\quad philippe.r.richard@umontreal.ca}
\and
Michel Gagnon
\institute{\'Ecole Polytechnique de Montr\'eal\\ Montr\'eal, Qc, Canada}
\email{michel.gagnon@polymtl.ca}}
\def\titlerunning{Improving QED-Tutrix by Automating the Generation of Proofs}
\def\authorrunning{L. Font, P. R. Richard, \& M. Gagnon}

\begin{document}
\maketitle

\begin{abstract}
The idea of assisting teachers with technological tools is not new. Mathematics in general, and geometry in particular, provide interesting challenges when developing educative softwares, both in the education and  computer science aspects. QED-Tutrix is an intelligent tutor for geometry offering an interface to help high school students in the resolution of demonstration problems. It focuses on specific goals: 1) to allow the student to freely explore the problem and its figure, 2) to accept proofs elements in any order, 3) to handle a variety of proofs, which can be customized by the teacher, and 4) to be able to help the student at any step of the resolution of the problem, if the need arises. The software is also independent from the intervention of the teacher. QED-Tutrix offers an interesting approach to geometry education, but is currently crippled by the lengthiness of the process of implementing new problems, a task that must still be done manually. Therefore, one of the main focuses of the QED-Tutrix' research team is to ease the implementation of new problems, by automating the tedious step of finding all possible proofs for a given problem. This automation must follow fundamental constraints in order to create problems compatible with QED-Tutrix: 1) readability of the proofs, 2) accessibility at a high school level, and 3) possibility for the teacher to modify the parameters defining the ``acceptability'' of a proof. We present in this paper the result of our preliminary exploration of possible avenues for this task. Automated theorem proving in geometry is a widely studied subject, and various provers exist. However, our constraints are quite specific and some adaptation would be required to use an existing prover. We have therefore implemented a prototype of automated prover to suit our needs. The future goal is to compare performances and usability in our specific use-case between the existing provers and our implementation.
\end{abstract}

\section{Introduction}

Geometry is a delicate subject to teach. One could believe that, since it is a science, a student will learn solely by accepting the concepts and following the processes of the existing theoretical model, with no regards for the reality of the world of space and shapes. By also accepting mathematics as an activity, we can consider the teaching of geometry as the development of an intuition about geometry, by the interaction between the theoretical model and the reality. In other words, ``doing geometry'' means working inside the theoretical model, by accepting all the codes and axioms, but ``learning geometry'' means developing this geometrical intuition of the way the theoretical model represents reality. In the light of this consideration, solving a demonstration problem, with or without technological tools, both develops the geometrical sense of the student and allows him to do his mathematician's work, which are the ultimate goal of mathematics education.

\begin{figure}
    \centering
    	\includegraphics[width=0.8\textwidth]{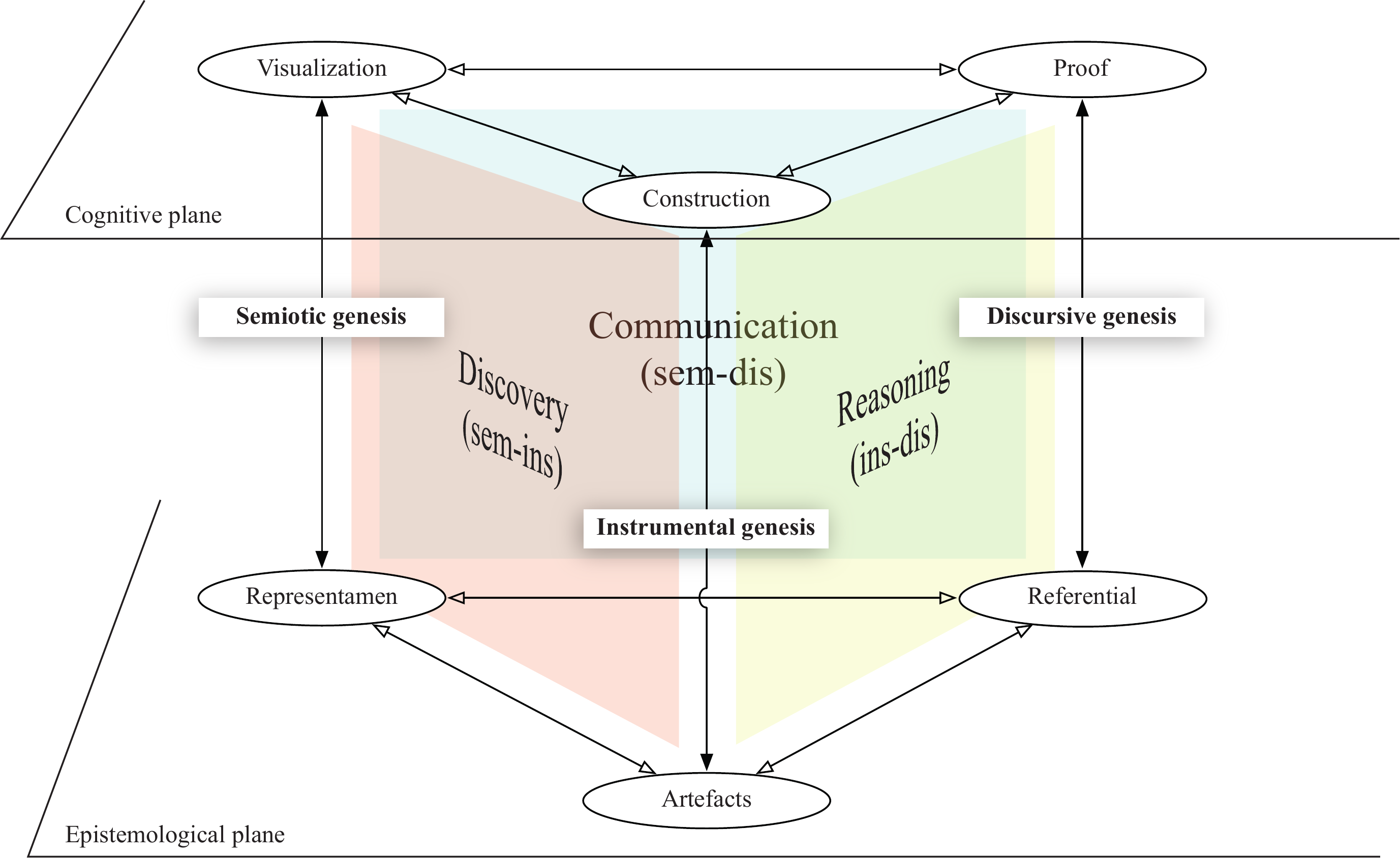}
	\caption{The model of mathematical working space in the theory.}
	\label{fig:etm}
\end{figure}

The theory of Mathematical Working Spaces (MWS) considers the outset that mathematics is both a science and an activity \cite{kuzniak2014espacios}. In a learning context, when the student does his mathematician's work, this dual perspective allows to interpret the development or the manifestation of geometric sense following \textbf{three geneses} at play between the epistemological and cognitive plans (see Fig.~\ref{fig:etm}): \textbf{instrumental}, \textbf{semiotic} and \textbf{discursive}. In a few words, the first one represents the familiarization with the tools, either ruler and compass, or software, and the use of them. The second one represents the association between mathematical concepts and processes, and their formal or systemic representation, such as symbols, semiotic representation systems and the technical mathematical lexicon. The third one represents the articulation of mathematical concepts and processes to form a proof, as the reasoning or calculations that proceed by discursive-graphic expansion \cite{richard2004inference}. These geneses allow us to classify crucial points, such as the creation of meaning, validation of properties, or usage of technological tools, in terms of genesis coordination in the model of MWS. For instance, a student solving a demonstration problem using software has, at a point, to identify the hypotheses and conclusion of the problem, including 1) understanding how the demonstration is articulated around them (discursive aspect), 2) finding the terms of statements carrying mathematical meaning (semiotic aspect), and 3) understanding how to manipulate, transform and control these mathematical elements through the software (instrumental aspect). 

For the intelligent tutor software developer who wants to integrate users (students or teachers) very early in the design process, the question of respect for human learning is a challenge at all times. When planning the possible interactions between the user and the machine, the developer has to know beforehand the hypotheses and conclusion, that are already part of a well-constructed, mechanically explorable proof. As a consequence, when the goal is to maximize the educational value of the software, it is necessary to introduce from the very beginning some cooperation between the experts of different domains (computer science, education science, cognitive science, user interface, etc.) by following collaboration-oriented research methods.

This requirement of integrating users as soon as the software is designed is crucial. Furthermore, in the theory of mathematics education situations \cite{brousseau2006theory}, and most specifically considering the notion of \textbf{didactical contract}, we consider that both student and teacher have reciprocal responsibilities regarding the knowledge devolution. The understanding of these various responsabilities is  fundamental for the development of an educational software. There are some explicit rules for knowledge sharing, such as a precise way of writing mathematics, the propriety used to justify the passage from one expression to another, etc., but most are implicit, being more a consequence of a class habit than a necessary constraint. Typically, in a deductive logic, the teacher can accept several inferential shortcuts, by accepting demonstrations that are not completely rigorous, in order to smoothen the flow of the resolution or improve its pedagogic efficiency. However, by doing so, he changes the operational structure based on definitions and properties in a given axiomatic. Besides, the student will build his own shortcuts while he gets used to the mathematics deductive paradigm. Both the question of the readability of proofs and the acceptability of an inferential shortcut are closely related to the didactical contract, i.e. mathematics as an activity. 

In this paper, we present our work on the intelligent tutor software, QED-Tutrix or QEDX (abbreviated acronym), that aims at providing a technological tool to improve the learning of geometry in highschool, by offering an interface to solve problems of proof while staying closely attached to the didactical contract. More specifically, we present the problematic of the \textbf{readability} and \textbf{accessibility} of proofs, and the \textbf{adaptability} of proofs to inferential shortcuts. These three constraints are the basis for the development of a system to automatically generate proofs, therefore easing considerably the task of adding new problems to QED-Tutrix. 

In Section~\ref{sec:relatedWork}, we present systems similar to QED-Tutrix, and also existing tools for automated theorem proving that could be useful for the task of generating proofs. In Section~\ref{sec:Presentation}, we provide a brief explanation of the functionalities of QED-Tutrix and its internal handling of proofs. Then, in Section~\ref{sec:goals}, we present our goals for the improvement of QEDX and the needs for such improvements, more specifically our need for an automated proof generator. We give a quick overview of the results of our preliminary work on a custom proof generator in Section~\ref{sec:deductive-engine}, and discuss the limits of our work in Section~\ref{sec:limitations}. Finally, we present other avenues of research and conclude in Section~\ref{sec:conclusion}.

\section{Related Work}
\label{sec:relatedWork}

This paper presents a project of improvement of an existing software, QED-Tutrix, in its capacity as an intelligent tutoring system, which allows the student to do his work as a mathematician when solving problems of proof in geometry. As a consequence, most of the related work has already been discussed in great detail in the theses of Nicolas Leduc \cite{Leduc2016} and Michèle Tessier-Baillargeon \cite{Tessier-Baillargeon2015}. In Section~\ref{subsec:rw-tutorSoftwares}, we summarize the important elements and, in Section~\ref{subsec:rw-theoremProving}, discuss the state-of-the-art for the novel problem of automated deduction.

\subsection{Tutor softwares}
\label{subsec:rw-tutorSoftwares}

In \cite{Leduc2016}, Nicolas Leduc analyzed the existing solutions for learning mathematics. He first identified non-tutor systems, divided in four groups: tools for autonomous learning; tools modeling the learning path and curriculum; micro-worlds; and tools for automated proving.

\textbf{Tools for autonomous learning,} typically websites providing answers to specific questions, such as Mathway \cite{mathway} or private tutoring companies. These are a fantastic knowledge bases, but are either passive tools or non-automated.

\textbf{Learner modeling,} the student is guided on a learning path and his knowledge is taken into account when giving him new content. Examples include one of the first such systems, ELM-ART \cite{weber2001elm} for learning LISP, and ALEKS \cite{falmagne2006assessment}, ActiveMath \cite{melis2009activemath} and Wayang Outpost \cite{arroyo2004web} for mathematics specifically. These require a modelization of the user's knowledge, that can be done using various techniques, such as fuzzy logic \cite{jean2000pepite} or Knowledge Components \cite{aleven2013knowledge}. These systems are based solely on the problems solved, and not on the way the problem was solved, which is one of the foci of QED-Tutrix.

\textbf{Micro-worlds,} the student, unlike in the paper-pencil environment, can manipulate a dynamic geometrical figure, following certain rules, such as Euclide's axioms. A typical example is GeoGebra \cite{hohenwarter2013geogebra}, a popular dynamic geometry software with an open source code and an active community. These tools are usually dependent on the presence of the teacher, since they offer little to no control over the student's actions, and no help towards the resolution of a problem.

\textbf{Automated proof,} these systems allow to verify statements, or discover new facts. These systems are discussed in detail in Section~\ref{subsec:rw-theoremProving}.

\paragraph{}
In a second part, he analyzed in detail 10 tutoring systems for geometry. Each of these systems offers interesting characteristics, but none combines them all. The goal for QEDX is to offer:
\begin{itemize}
\item an interface to explore the problem by allowing the student to freely manipulate the figure;
\item a liberty in the construction of the proof, allowing the student to construct his proof in any order;
\item a handling of all possible proofs, acceptable at a high-school level;
\item a tutor system to help the student, based on the identification of the step of the proof on which he is working;
\item autonomy from the teacher, allowing an unsupervised use by the students.
\end{itemize}

In the remaining of this section, we provide a short summary of the systems analyzed and explain their shortcomings.

One of the first tutor system developed for geometry is GeometryTutor. It is based on a solid theoretical model, the ACT-R cognitive theory \cite{anderson1996act, anderson2000implications}. However, it does not allow the student to explore the problem outside of the rigorous path identified by the software. 

A few years later, the PACT Geometry Tutor has been developed \cite{aleven1998combatting, aleven2000limitations}, that evolved into Geometry Cognitive Tutor \cite{aleven2002effective, aleven2006toward, roll2014benefits}. This system is limited to problems based in cartesian coordinates, since it handles elements numerically. In QEDX, we want to be able to handle any geometry problem.

To include proofs that require an additional construction, the Advanced Geometry tutor was developed by Matsuda \cite{matsuda2005advanced}, based on the GRAMY theorem prover \cite{matsuda2004gramy}. This system has interesting characteristics, since it is one of the few to handle proofs with intermediate constructions. However, it is quite rigid on the accepted proof and does not allow the student to explore the problem or use a proof that is not the optimal proof calculated by the prover.

The software ANGLE \cite{koedinger1991design,koedinger1993effective}, unlike the previous ones, aims at helping the student to construct his proof. It is based on the Diagram Configurations \cite{koedinger1990abstract} theory, based on interesting configurations of the geometrical figure, used by the experts to produce the proof. It gives the student freedom to explore the solutions. However, the Diagram Configurations are modeled on the work of experts, which can be quite far from the formal geometry taught in highschool. Besides, even though the student can explore the problem, he has no interface to manipulate the figure.

An interesting approach is the Baghera system \cite{balacheff2003baghera,webber2001baghera}, providing a web platform. This system offers two sides: one for the teacher, where he can create new problems and follow in real-time or replay the progress of the student; and one for the students, which can chose and solve problems. The weakness of this system when compared to our goals is that it offers no automated tutor.

To provide interactive figure manipulation, it is a logical step to use an interactive geometry software. Two systems are based on the Cabri software \cite{baulac1990micromonde, kordaki2006potential}, Cabri-DEFI \cite{luengo1998contraintes} and Cabri-Euclide \cite{luengo1997cabri,luengo2005some}. The first one helps the student to plan his proof by asking him questions about the figure that ultimately direct him towards a proof. It is an interesting approach, but it offers no freedom of exploration, since it is the system that asks questions. Besides, there is no tutor system to help the student when he is stuck in his resolution. The second one goes in the opposite direction, by allowing the student to explore freely and enter conjectures, that are later organized in a graph. However, there is no help provided to the student and no mechanism to ensure that the problem is ultimately solved, which can be an issue for unsupervised use.

The system Mentoniezh \cite{py1994reconnaissance,py1996aide,py2001environnements} offers a novel approach by dividing the proof in four steps: understanding the problem; exploration of the figure; planing of the proof; and redaction. The software helps and directs the student during each step. The division of the proof in steps is one of the foundation of QEDX. However, the software provides no help to find the next proof element, and a student can therefore encounter an impasse in his resolution that will force him to ask the teacher for help.

Another system dividing the proof in steps is Geometrix \cite{geometrix}, where the student can first construct a figure, and then allows him to create a problem based on that figure and to solve it. It therefore allows the creation of exercises by the teachers, including customized error messages to help the student. However, it remains mainly a demonstration assistant, and offers little in the tutoring aspect. 

Finally, the Turing system \cite{el2005development,richard2007amelioration}, that largely inspired QEDX, provides an interface for the student that allows him to manipulate a dynamic figure, and to provide statements to construct his proof, in any order. The integrated tutor system analyzes his input and gives feedback depending on the validity of the statement. After a period of inactivity, the tutor gives him hint to restart his resolution process. However, the hint is based only on the last element provided by the student, which may not be the step on which he is currently working.

Overall, all of these systems, except ANGLE, are based on fully formal geometry, which limits the number of acceptable proofs, even though less formal proofs are typically accepted by the teachers. ANGLE is based on a model of the reasoning of experts, which is quite far from the proofs used in class. Besides, only Mentoniezh keeps the previous work of the student in memory, but does not use it to provide hints towards the next step. The systems that provide hints are based on forward or back-tracking, limiting their usefulness. Finally, no system allows the student to explore different proof paths at the same time. This illustrates the need that gave birth to the QEDX project.

\subsection{Automated theorem proving}
\label{subsec:rw-theoremProving}

The main issue faced currently by QED-Tutrix is the difficulty of the task of adding new problems. Indeed, to help the student in his resolution, one must know the element(s) of the proof on which he's working, which requires a knowledge of all possible proofs accessible for the student at a given point in the school curriculum. Therefore, the software must internally handle a representation of these. This information is stored in the form of inference graphs (see Section~\ref{subsec:InternalEngine}). These graphs can be quite large even for simple problems, and we currently have to construct them manually, hence the need for a tool to automatically find all possible proofs. However, we have three fundamental constraints:
\begin{itemize}
\item the proofs must be \textbf{readable};
\item they must use only properties available at a \textbf{high-school level};
\item there must be a way to handle the \textbf{inferential shortcuts}, i.e. the inference chains that can be deemed too formal by some teachers and are therefore skipped in a demonstration.
\end{itemize}
These three points are detailed in Section~\ref{subsec:pre-requisites}, along with an example of an inferential shortcut.

The constraints direct our search for a way to automatically find proofs. Indeed, there currently exist two general research avenues for geometry automated theorem provers (GATP): algebraic methods and synthetic, or axiomatic, methods. The first one is based on a translation of the problem into some form of algebraic resolution, and the second one uses an approach closer to the natural, human way of solving problems, by chaining inferences.

One of the main goals of the research community in automated theorem proving is the performance. Since synthetic approaches are typically slower, most solvers are based on an algebraic resolution. Algebraic methods include the application of Gröbner bases \cite{buchberger1988applications,kapur1986using}, Wu's method \cite{chou1988introduction,wu1979elementary} and the exact check method \cite{zhang1990parallel}. Practical applications include the recent integration of a deduction engine in GeoGebra \cite{botana2015automated}, which is based on the internal representation of geometrical elements in complex numbers inside GeoGebra. Other examples include the systems based on the area method \cite{boutry2016tarski,janicic2006gclc}, the full-angle method \cite{chou1994machine,chou1996automated}, and many others. These systems seldom provide readable proofs, and when they do, they are far from what a high-school student would write. Given our readability and accessibility goals, all these systems are not relevant to our interests.

For this reason, we focus on synthetic methods. A popular approach is to use Tarski's axioms, which have interesting computational properties \cite{braun2017synthetic,narboux2006mechanical}. However, the geometry taught in high-school is based on Euclide's axioms, which are not trivially correlated to Tarski's. Therefore, proofs based on Tarski's axioms are quite inaccessible for high-school students, violating our second constraint. 

A prover that has very similar goals is GRAMY \cite{matsuda2004gramy}. It is based solely on Euclidean geometry, with an emphasis on the readability of proofs. Besides, it has been developed as a tool for the Advanced Geometry Tutor, that has been presented in Section~\ref{subsec:rw-tutorSoftwares}. It is therefore able to generate all proofs for the given problem. Finally, one of its major strengths is the ability to construct geometrical elements. 

To the best of our knowledge, GRAMY is the only work close to our goals. It is therefore one of the basis for our work on automated generation of proofs. The only limit to the compatibility of the two systems is our third constraint: the axioms used by our system must be flexible and easily changed by the teacher to suit his personal teaching style and to follow the evolution of the axioms taught in class throughout the year.

Overall, given our very specific needs dictated by the focus on educational interest, we identified only one system, GRAMY, that would be suitable. However, since its code is closed and no work has been done on it since 2004, we chose to develop a custom engine inspired by it. This will allow us to integrate it directly to QED-Tutrix.

\section{Presentation of QED-Tutrix}
\label{sec:Presentation}

\begin{figure}
    \centering
    	\includegraphics[width=0.5\textwidth]{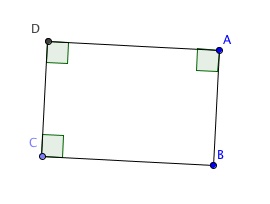}
        
    	 \textit{``Prove that any quadrilateral with three right angles is a rectangle.''}
	\caption{An example of a simple geometrical proof problem.}
	\label{fig:proofProblemExample}
\end{figure}

The QED-Tutrix software, or QEDX, has benefited from the technological and scientific achievements of its precursors, notably the geogebraTUTOR and Turing systems. It aims at providing an interface in which the students can solve geometrical proof problems, such as the one in Fig.~\ref{fig:proofProblemExample}.

It is first and foremost a research project, with the central goal of adapting the technology to the student, instead of adapting the student to the technology. This project is therefore conducted by a close cooperation between researchers in mathematics education and in information technology, and every step of its development included a profound reflexion on the educational value of the software.

\subsection{Conception guidelines}
\label{subsec:ConceptionGuidelines}

To ensure the educational interest, we attempted to follow several guidelines. The most general one is to adapt the software to a typical student's thought process. In other words, to reduce as much as possible the work that has to be done to understand and use QEDX. This way, it becomes a tool that can be used almost intuitively to write proofs. This required several test sessions in class, during which the behavior and reactions of the students were closely monitored. These studies are explained in detail in the work of Michèle Tessier-Baillargeon \cite{Tessier-Baillargeon2015} . 

Since QEDX is a software for solving proof problems, we based our work on the way a typical student solves such problems. Usually, the resolution of a proof problem implies three steps:
\begin{itemize}
	\item \textbf{Exploration}, during which the student gets acquainted to the problem and searches for ideas, without a plan on the exact proof he is going to write;
    \item \textbf{Construction} of the proof, during which the student finds out which elements of the class (theorems or lemmas, for example) can be used to materialize his ideas;
    \item \textbf{Redaction} of the proof, during which the student organizes his ideas to write a formal proof.    
\end{itemize}
These steps are not usually followed linearly. A student will often try a proof, to realize during its redaction that it does not work, or that an inference is missing somewhere, sending him back to the exploration. 

\begin{figure}
    \centering
    	\includegraphics[width=0.9\textwidth]{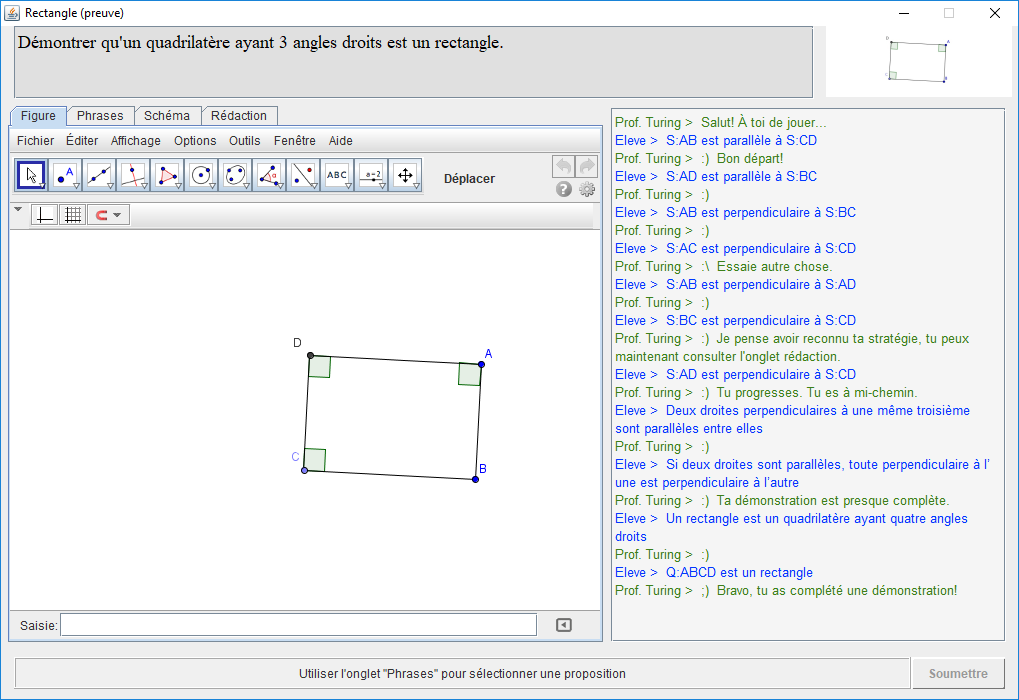}
    \caption{The main interface of QED-Tutrix.}
	\label{fig:interface-figure}
\end{figure}

\begin{figure}
    \centering
    	\includegraphics[width=0.9\textwidth]{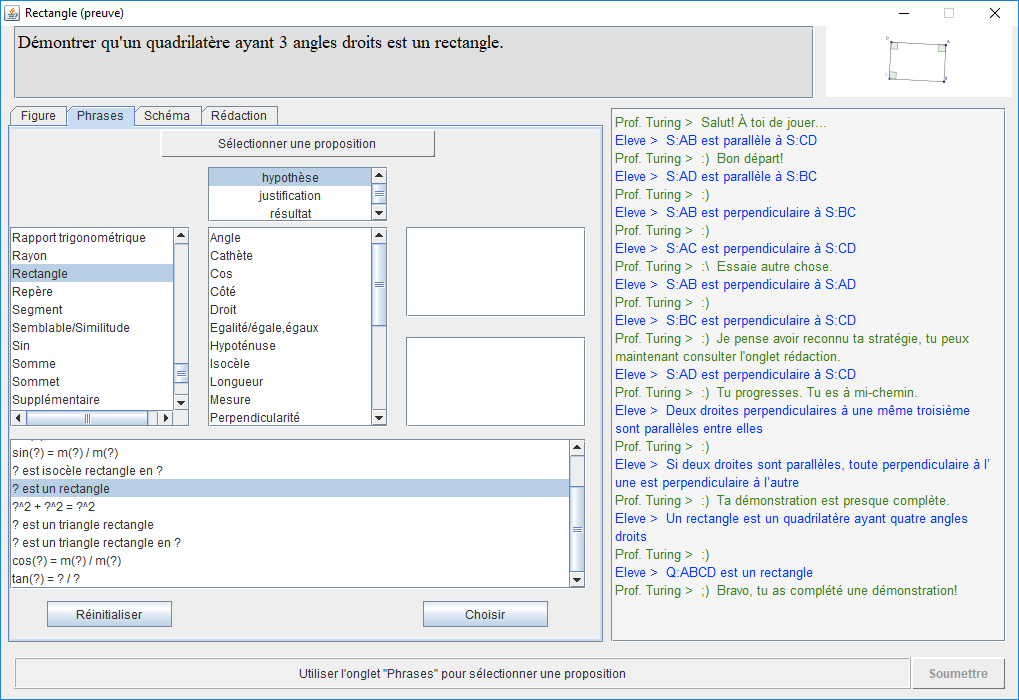}
    \caption{The main interface of QED-Tutrix, on the ``Sentence'' tab.}
	\label{fig:interface-phrases}
\end{figure}

\begin{figure}
    \centering
    	\includegraphics[width=0.9\textwidth]{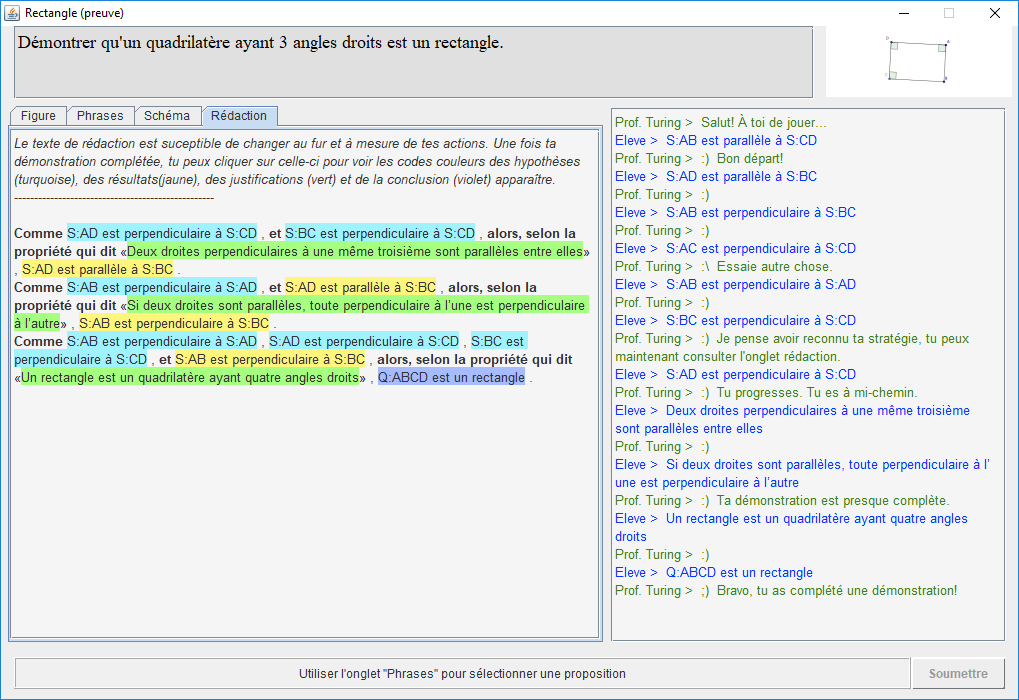}
    \caption{The main interface of QED-Tutrix, on the ``Redaction'' tab.}
	\label{fig:interface-redaction}
\end{figure}

The interface of QED-Tutrix, represented in Fig.~\ref{fig:interface-figure} reflects this reality. It is composed of four sections:
\begin{itemize}
\item The topmost is simply the statement of the problem, including a figure. 
\item The left is the core of the communication from the student to QEDX. It is composed of three tabs, plus one currently not implemented.
\item The right is the text box containing the artificial conversation going on between the tutor, embodied by Prof. Turing, and the student. We explain the role of this conversation later in this section.
\item The bottom is the input interface, where the student can complete the statement he wants to put in his proof.
\end{itemize} 

The three functional tabs of the left section are associated with the three steps for writing a proof. Indeed, the first tab, ``Figure'', is the one present in Fig.~\ref{fig:interface-figure}. It uses a GeoGebra plugin that allows the student to freely manipulate the figure, for example by measuring angles or adding lines, and therefore \textbf{exploring} the problem. GeoGebra is a popular tool for dynamic geometry that contains many functionalities, and is therefore an excellent way for the student to experiment. Besides, its popularity ensures that many students will be familiar with it.

\begin{figure}
\centering
    	\includegraphics[width=0.9\textwidth]{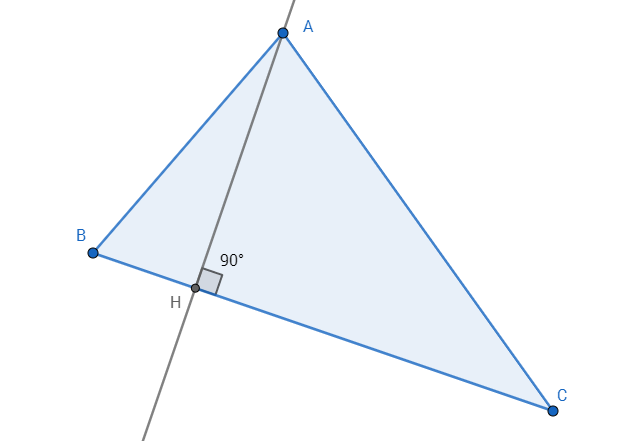}
    \caption{A geometrical construction of the height of a triangle.}
	\label{fig:triangle-height}
\end{figure}

The second tab, ``Sentences'' (``Phrases'' in French), shown in Fig.~\ref{fig:interface-phrases}, allows the student to \textbf{construct} his proof by selecting mathematical statements or properties. This selection is done by four narrowing lists of mathematical topics: the first one is fixed and contains all the topics accessible to the student, then the second, third and fourth adapt to the choice made in the first one by narrowing the options. At every step, the bottom window contains a list of statements compatible with the selected topics. When his choice is made, the bottom of the interface allows him to choose the geometrical element(s) concerned. For instance, in the situation depicted in Fig.~\ref{fig:triangle-height}, if he wants to indicate the result ``AH is the height of triangle ABC through A'', he can start by selecting ``Triangle'' in the first list of topics. Then, in the second list, he selects ``Height''. This combination of topics allows the statement ``? is the height of ? through ?'' to appear in the bottom list. He clicks on it, then fills the fields that appears at the bottom with ``AH'', ``ABC'' and ``A''.

Finally, the third tab, ``Redaction'', shown in Fig.~\ref{fig:interface-redaction}, provides a fully written rigorous proof. This tab is accessible only when the software is able to determine that the student has provided enough elements (arbitrarily fixed at 50\%) of any proof. The full formal proof, however, contains only the elements already entered by the student: the remaining ones are blanks. This helps the student in the \textbf{redaction} of his proof, by directing him towards the missing elements.

Overall, the typical usage of our software consists in a constant navigation between the tabs, switching between exploration, construction and redaction. In this aspect, QED-Tutrix embraces the reasoning of its user.

\paragraph{}
A crucial element of the software, that has not been covered yet, is the tutor interface. Indeed, the goal of QEDX is not simply to provide an interface to solve problems, but also to be an intelligent tutor, helping the student throughout his resolution of the problem.

Our tutoring system works by finding out the proof the student is trying to write. The software has an internal representation of all possible proofs for a given problem, and finding the one is simply a question of calculating the proof for which the student has entered the most elements. More details on the internal representation and exploration of proofs are given in the following section.

This calculation is done dynamically, every time the student enters a new element into the software. This way, when the student reaches an impasse in his resolution, the software identifies an element of his proof that is yet missing and directs him towards it by printing a customized message in the right panel. An example of such a message could be:
\begin{center}
    \textit{``What are the definitions of the height of a triangle ?''}
\end{center}
If the student does not find anything after several advices, the software tries to direct him towards another missing element. If this is still not enough to exit the impasse, we consider that the student is either completely stuck or has become distracted, and the tutor asks the student to consult his teacher.

The interest of this system is to allow all students of a class to obtain a personalized help during the entirety of the problem resolution, compared to a typical class format where the teacher cannot physically help all his students at the same time. Even though the help provided by the software is more crude and mechanical than the one provided by a teacher, that has experience and intuition, the combination of both seems like an excellent way to help the students in their learning process.
    
\subsection{Internal engine}
\label{subsec:InternalEngine}

To allow this versatility, we introduced some novel elements in QEDX' core. First, for it to run on as many system as possible, QED-Tutrix is coded in Java. Furthermore, it allowed us to use the GeoGebra API \cite{geogebraAPI} to provide our exploration interface very simply. 

\begin{figure}
    \centering
    	\includegraphics[width=1\textwidth]{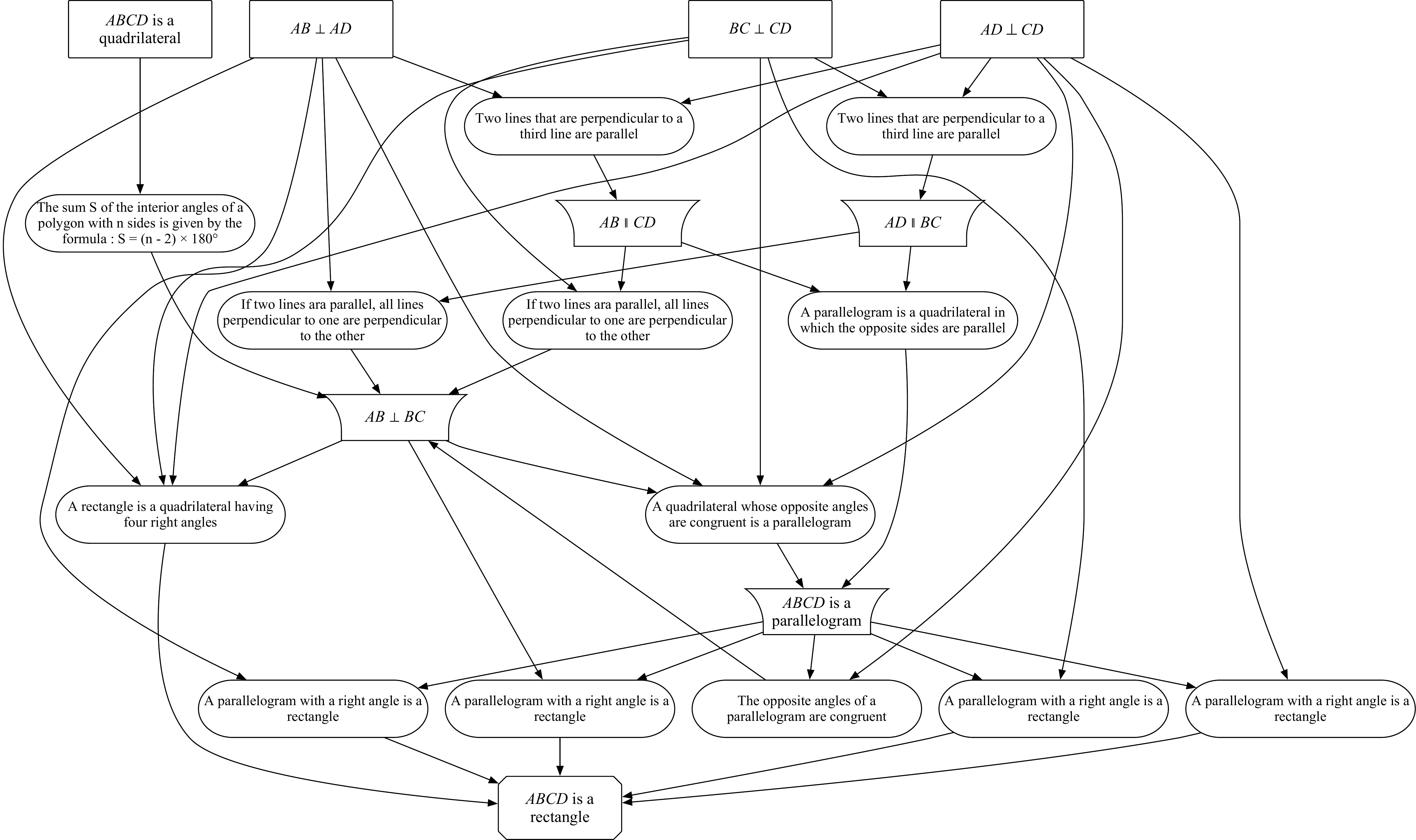}
    \caption{The HPDIC graph for the rectangle problem.}
	\label{fig:HPDIC-rectangle}
\end{figure}

A central element of our engine is the \textbf{HPDIC graph}, standing for Hypothesis, Property, Deduction, Intermediate result, Conclusion. In a few words, this graph contains all possible proofs for a given problem, in a set of axioms. This set of proofs evolves as the students can (and should) use more advanced properties, opening the possibility for more diverse proofs. We therefore consider by default the maximum graph, or \textbf{complete graph}, containing all proofs that could be used at any point in the high-school curriculum. If we see a proof as a succession of inferences (hypotheses + property $\rightarrow$ conclusion), then a proof is a directed graph, whose starting nodes are the hypotheses, and finishing node the conclusion. Then, to represent all the proofs of a problem, we fuse these into a larger graph. For example, given the problem in Fig.~\ref{fig:proofProblemExample}, \textit{``Prove that any quadrilateral with three right angles is a rectangle''}, its graph would be the one in Fig.~\ref{fig:HPDIC-rectangle}.  In such a graph, a proof is a subgraph in which every ``intermediate result'' node has only one parent, i.e. one justification. This figure is taken from a previous paper \cite{richard2017connectedness}, where we first discussed the possibility of including connected problems in QEDX.

During the resolution of a problem by a student, every input, either statement or property, must be present on the graph, otherwise we consider that it is not part of the proof. At the beginning of the resolution, all nodes of the graph are ``unchecked''. When the student adds an element, it is ``checked'' on the graph. This way, identifying the proof he is working on is a matter of checking which subgraph has the highest proportion of checked nodes.

\begin{figure}
    \centering
    	\includegraphics[width=0.5\textwidth]{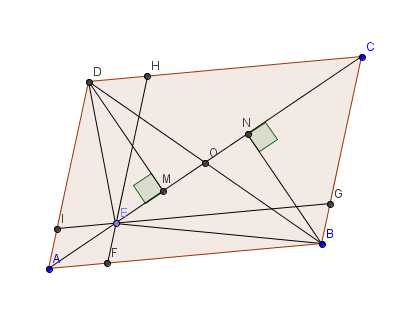}
        
         \textit{``If E is a point of the diagonal AC of parallelogram ABCD, what is the relation between the areas of triangles AEB and AED ?''}
    \caption{The parallelogram area problem.}
	\label{fig:problem-parallelogram}
\end{figure}

This check, however, must be done by exploring the space of all possible proofs, i.e. of all subgraphs from the hypotheses to the conclusion. In the rectangle problem, whose graph is represented in Fig.~\ref{fig:HPDIC-rectangle}, this number remains small. However, this problem has an unusually simple graph. In another problem that is not much more complicated, shown in Fig.~\ref{fig:problem-parallelogram}, the HPDIC graph has several thousands nodes, for over five million possible solutions. This makes the search of subgraphs impossible in real-time, especially on computers typically used in class. Therefore, to avoid this issue, the graph is converted into a series of tree. This operation also simplifies the creation of the space of possible proofs by removing cycles. The exact process is described in the thesis of Nicolas Leduc \cite{Leduc2016}. This transformation is lengthy, but only has to be done once, and allows the software to explore the set of trees in real-time, rendering it possible to find, at every instant, the proof the student is likely working on, and to provide an instant feedback.

\section{Goals for the improvement of QED-Tutrix}
\label{sec:goals}

Even though QED-Tutrix is currently functional and already provides interesting educative features, we identified many improvement avenues. The most research-oriented one is drawn from the following observation: when a student is in an impasse in a problem resolution, most teachers help the student towards the missing steps of the problem; however, some teachers chose an alternative approach by giving the student another problem, \textbf{connected} to the original one. QEDX currently mimics only the first approach, and one of our long-term goals would be to also provide the second approach, by providing connected problems in impasse situations.

More specifically, we consider that two problems are \textbf{connected} if their proofs use at least some common mathematical elements. For instance, two problems using the Pythagorean theorem are connected, whereas a third problem about the properties of the hexagon is not connected to the first two. There is no order relation, the problems can be of equal difficulty and length, or very different. However, in this context, they should be approximately close in difficulty, since there is limited educational interest in helping a student stuck on a simple problem by asking him a difficult one. The \textbf{connectedness} of two problems is their similarity, a low connectedness means that two problems have few elements in common, whereas a high connectedness means that they are almost similar. The \textbf{difficulty} of a problem is not trivial to establish. Providing a rigorous definition to represent our intuition will be one of the very next steps of our work on connected problems.

A previous paper by Fortuny, Gagnon and Richard \cite{richard2017connectedness} illustrates the educational interest of this approach, both for the student's learning and for the mathematics education research community. In class, the submission of a new, connected problem to the student can restart a halted solving process. Our goal is to mimic this behavior. The main difficulty is that the teacher choses the new problem almost instinctively, helped by his experience. It is therefore a difficult task, even for a teacher, to identify the exact criteria used to determine, first, if giving another problem instead of clues towards the solution would be profitable, and second, \textbf{which} problem to give in every particular situation. The first point could be tackled by experimenting with both approaches in class. Concerning the second point, our intuition is that it essentially boils down to identifying the exact step of the proof the student is stuck, which is exactly one of the functionalities of QEDX, and then find another problem using a similar proof. Of course, before working on the connected problems functionality, this intuition will have to be thoroughly verified. This, while interesting, still remains a goal for the future.

Indeed, a fundamental issue arises when trying to tackle the connected problems approach. As we mentioned previously, the HPDIC graph, that must be provided when implementing a problem in QEDX, could be huge. Currently, we have \textbf{no method} to automatically generate such a graph, and it therefore has to be constructed manually, which is a tedious and lengthy operation. For this reason, QEDX only contains five problems in its current state, but an approach such as the connected problems one requires a consequent number of available problems covering each topic seen in class. Five is not nearly enough, especially since we chose those to be as different as possible, in order to cover various topics and proof types with few problems.

Therefore, a necessary step, both for the implementation of connected problems and for the usability of QEDX in general, is to pipeline the production of new problems. This includes two steps: identifying interesting problems, which can easily be done by searching the geometry textbooks, and implementing them in QEDX, which includes the difficult step of writing down all possible proofs. Among the research avenues for the improvement of QEDX (see Section~\ref{sec:conclusion}), this paper specifically presents the automated generation of proofs, in order to supply QED-Tutrix with a healthy amount of problems.

\section{Automated generation of proofs}
\label{sec:deductive-engine}

The idea of proving geometry theorem automatically is not new. As soon as 1960, Gelernter \cite{gelernter1960empirical} proposed a synthetic proof method for geometry theorem proving. Other early approaches include those by Nevins \cite{nevins1975plane}, Elcock \cite{elcock1977representation}, Greeno et al. \cite{greeno1979constructions}, Coelho and Pereira \cite{coelho1986automated}, and Chou, Gao and Zhang \cite{chou1993automated}. Since then, the methods have evolved considerably. However, our use-case is very specific for several reasons.

\subsection{Pre-requisites}
\label{subsec:pre-requisites}

First, our usage of proofs is to spot the position of a student in a graph of proofs during his resolution process. Therefore, the only proofs that are relevant are proofs that could have been written by a high-school student. We consider a proof to be \textbf{readable} if it only uses properties available at a high-school level, and if its length, i.e. the number of inferences, is not unreasonably high. This definition is still informal and subjective, and we have yet to reach a consensus, internally and with external teachers, on what exactly is a readable proof. We discuss part of this issue in Section~\ref{sec:conclusion}. Still, even without a rigorous definition, this criteria already eliminates all tools based on an algebraic methods, since those provide proofs that are very far from classic Euclidean geometry.

Second, we mentioned that a proof must be accessible to a high-school student. This criteria changes throughout high-school classes, only some properties are available at first, and, as the class advances, more properties are studied, and, therefore, available for the students to use. Hence our second criteria, that the set of properties on which the inference engine works must be fully customizable by non-experts, typically teachers, who will adjust the available properties as the school-year advances.

\begin{figure}
    \centering
    	\includegraphics[width=0.5\textwidth]{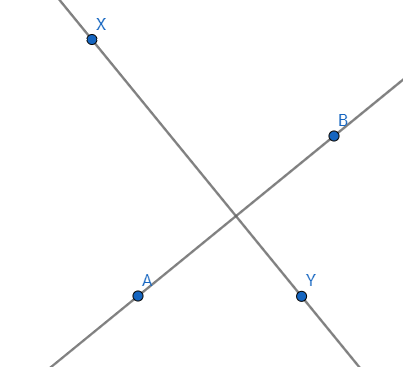}
        
         \textit{``X and Y are equidistant from A and B. Prove that (XY) is the perpendicular bisector of [AB] ''}
    \caption{A perpendicular bisector problem.}
	\label{fig:perp-bisector}
\end{figure}

Finally, teachers are all different in their teaching methods. More specifically, when they ask students to solve a proof problem, they don't require the same level of rigor in the student's proof. Hence the notion of \textbf{deductive isle}. This notion has been defined in \cite{richard2017connectedness}: ``\textit{Deductive isles} is our translation from the French \textit{îlot déductif} that considers the network of mathematical properties and definitions accepted or actually used in a given class, which includes the implicit hypothesis and the inferential shortcuts tolerated in the didactic contract.'' For example, in the situation described in Fig.~\ref{fig:perp-bisector}, most teachers would accept the following proof:
\textit{\begin{enumerate}
\item Every point on the perpendicular bisector of a segment [AB] is equidistant from A and B.
\item Since X and Y are both equidistant from A and B, the line (XY) is the perpendicular bisector of [AB].
\end{enumerate}}

However, some teachers would not consider it as rigorously valid. Instead, they expect this:
\textit{\begin{enumerate}
\item Every point on the perpendicular bisector of a segment [AB] is equidistant from A and B.
\item Since X is equidistant from A and B, X is on the perpendicular bisector of [AB].
\item Same reasoning for Y.
\item Since X and Y are two distinct points \{1\}, there is one and only one line (XY) passing through X and Y.
\item Since X and Y are both on the perpendicular bisector of [AB], (XY) is the perpendicular bisector of [AB].
\end{enumerate}}

In this example, these two inference chains put together constitute a deductive isle, since they use a different granularity of properties to draw the same conclusion (the line (XY) is the perpendicular bisector of [AB]) from the same hypotheses ([AB] is a segment, and X and Y are equidistant from A and B).

The handling of, and the ability to parametrize these deductive isles is essential for the future development of QED-Tutrix. Indeed, we want teachers to be able to use QEDX as a projection of themselves, allowing them to provide a personalized help to twenty or thirty students at the same time.

These three points, readability of the proof, flexibility in the axioms, and handling of the deductive isles, are the fundamental criteria for the automated generation of proofs. 

\subsection{Our custom engine}
\label{subsec:custom-engine}

We initially had the option to adapt an existing theorem prover or inference engine to suit our needs. Given the fact that optimization and efficiency are not our goal, and that our constraints are quite specific, we decided to work on the implementation and integration of a light custom engine in QEDX, written in Prolog. Prolog is a well-suited language for this task, and the engine itself is quite straightforward to program. The difficulty arises in the encoding of problems. Indeed, if we look back at an extremely simple problem such as the one of the perpendicular bisector in Fig.~\ref{fig:perp-bisector}, we already have to add the implicit hypothesis that X and Y are distinct to write the formal proof. This is a typical example of an hypothesis that is obvious both for the students and the teachers, but not for a mechanical tool. Therefore, the hypotheses must be carefully enumerated for the engine to work.

Besides, a recurrent problem in geometry theorem proving is the \textbf{construction of intermediate elements}. For instance, many geometry problems require the construction by the student of a geometric element, such as the height of a triangle or the diameter of a circle. Knowing which element to construct is a difficult step for an automated prover. Since the main goal is not to develop a general and well-rounded engine, but instead to reduce the workload to add new problems to QED-Tutrix, we opted for an alternative solution, by imposing that the figure given as an input to the solver include all those intermediate elements. We define that figure as the \textbf{super-figure} of the problem, whereas the \textbf{figure} represents only the geometrical elements given to the student as part of the problem. This requisite is fundamental, since it defines the nature of the prover, it is not strictly speaking a reasoning tool, but only a mechanical tool. We provide details on the way it works later in this section. Since this research project is not an automated deduction project, we deemed this compromise acceptable for the time being. We keep in mind the possibility of exploring ways to automate this construction step in the future.

The operation of our engine is quite simple. It is composed of a Java handler, calling a Prolog reasoner iteratively, building new results at every step, until nothing remains to be inferred. The initial problem is defined by a set of Prolog statements, one for each hypothesis of the problem. ``Hypothesis'' in this context is quite broad, since we must define all objects. \texttt{point(a)} is a hypothesis, \texttt{line(a,b)} as well.

Then, Prolog explores all known properties to extract the ones that can be used with this set of hypotheses to infer new results. In the next iteration, the obtained results are added to the set of hypotheses usable by the properties. We continue iterating until no new inference can be made. The Java handler checks at every step if the result is already known before adding it to the set. The algorithm is detailed in Fig.~\ref{fig:algorithm-prolog}.

\begin{figure}
	\begin{algorithmic}[1]
    	\REQUIRE $problemFile.pl\quad exists$
        \REQUIRE $properties.pl\quad exists$
		\STATE $S, C \leftarrow problemFile.pl$
        \COMMENT{S for Statements, C for Conclusion}
        \STATE $P \leftarrow properties.pl$
        \COMMENT{P for Properties}
        \REPEAT 
        \STATE $R \leftarrow Reasoner(S,P)$
        \COMMENT{R for Results}
        \STATE $N \leftarrow R - S$
        \COMMENT{N for New results. ``-'' is the relative complement operation on two sets}
        \STATE $S \leftarrow S + N$
        \COMMENT{``+'' is the Union of two sets}
        \UNTIL{$IsEmpty(N)$}
        \IF{C in S}
        \STATE $G \leftarrow ConstructGraph(C, S)$
        \COMMENT{G for (HPDIC) Graph}
        \RETURN $G$
        \ELSE
        \PRINT $Error$
        \RETURN $0$
        \ENDIF
        
	\end{algorithmic}
    \caption{The inference engine algorithm.}
    \label{fig:algorithm-prolog}
\end{figure}

The result of this algorithm is, first, a set of all inferred statements, and, second, if the conclusion is among them, as it should be, the HPDIC graph for the problem. Therefore, the pipeline for the addition of a new problem in QEDX would be:
\begin{enumerate}
\item Creation of the problem, redaction of the hypotheses and conclusion;
\item Construction of the super-figure, i.e. all geometrical elements that are useful for at least one proof;
\item Translation of the super-figure and hypotheses into Prolog facts;
\item Generation of the HPDIC graph by the inference engine;
\item Integration of the new problem to QEDX.
\end{enumerate}
In this list, steps 1 to 3 must be done by humans, but the bulk of the workload is in step 4. Step 5 is a target for automation later in the project. Overall, the usage of an inference engine reduces drastically the time needed to create new problems for QEDX.

\section{Limitations}
\label{sec:limitations}

This project has a number of limitations. We categorized these in two categories: the limitations of the software itself as it is currently, and the limitations of the inference engine we are currently building.

\subsection{QED-Tutrix in its current state}
\label{subsec:limitations-current}

QED-Tutrix has globally been well received when tested in class during the experimentations conducted by Michèle Tessier-Baillargeon \cite{Tessier-Baillargeon2015}, following the precepts of \cite{richard2011didactic} and \textit{conception by usage} \cite{rabardel1995hommes}. Teachers and students both appreciated the value provided by its usage. However, we also collected several suggestions for possible improvements. 

First, the interface has been designed to be as close to the reasoning of the student as possible. Still, some elements are not yet as easy to use as we want them to be. Typically, the current ``sentences'' tab is not an ideal way for students to select statements and properties: it requires some search, the statements themselves are not always properly organized, and it is sometimes not easy to know where to search for an element. An ideal solution for this would be to implement a natural language parser, in order to allow the student to write directly his element, and then giving him a list of possible geometrical elements corresponding to his input. Another team is currently working on this possibility.

Second, as mentioned in Section~\ref{sec:goals}, there are only five problems that have been manually implemented in QEDX, with no way to add any significant number in a reasonable time. This issue is the result of the  difficulty and time consumption of finding and encoding all possible proofs by hand: these five problems are the result of three years of development, and have been chosen to be as diverse as possible. Still, until it is addressed, QEDX will remain a tool that cannot be used for real, long-term experiments. This important issue of the software is the reason behind the research project described in this paper.

\subsection{Our inference engine}
\label{subsec:limitations-engine}

One must keep in mind that this project is still in its infancy. The inference engine we started to build is currently only a proof of concept. It still is advanced enough to allow us to identify several limitations to our approach.

We mentioned in Section~\ref{subsec:custom-engine} that there can be no implicit hypothesis when using our inference engine. The hypotheses given in input to the inference engine must contain all the relevant information, such as the geometrical elements involved, including the intermediate elements not given to the student their relation, and other hypotheses not contained in the figure. This requires some manual work, including giving names to the objects. For example, the statement ``Prove that any quadrilateral with three right angles is a rectangle'' is not usable directly. A correct set of statements would be ``Given that $A,B,C$ and $D$ are distinct points,  that $AB, BC, CD$ and $AD$ lines, that $AB \perp BC$, $BC \perp CD$ and $AB \perp AD$, and that $ABCD$ is a quadrilateral, prove that $ABCD$ is a rectangle''. The redaction of completely rigorous hypotheses is a tedious task for the teacher that wants to add a problem to the software.

Furthermore, the necessity of providing the super-figure when using the engine, albeit as a temporary solution, limits the interest of our engine. Indeed, it transfers a great deal of the difficulty to the human. Still, on the five problems currently implemented in QEDX, only two need intermediate constructions, three can be solved completely without construction. We plan to analyze further the proportion of problems that need constructions among the problems accessible to a high-school student. If needed, we plan to explore solutions later in the project to automatically or semi-automatically generate these super-figures. The approach used in the prover GRAMY \cite{matsuda2004gramy} may be a possible solution.

\section{Other avenues and conclusion}
\label{sec:conclusion}
 
In this paper, we presented our plans for the improvement of QED-Tutrix by automatically generating proofs, allowing us to integrate new problems easily, and therefore increasing the range of the software. It is however not the sole research avenue. Indeed, several other aspects of QEDX offer room for improvement. Even when the generation of proofs will be fully automated, the task of encoding the hypotheses, including the implicit and figural ones, remains a tedious task, especially for a teacher who does not want to learn a meta-language. It would therefore be extremely interesting to be able to extract the formal hypotheses, explicit and implicit, and the conclusion, from the statement of the problem in natural language. This task is directly related to the semiotic genesis, i.e. to transform conceptual, abstract knowledge into formal elements, whereas the automated generation of proofs is in the discursive genesis. Another team in the QEDX research group is currently working on this project.

Another interesting aspect is the enhancement of the tutor system. Indeed, the detection of impasses is currently quite simple: the system provides a hint when the student has not given any new element in an arbitrarily fixed time. It would be interesting to improve this impasse detection, for instance by logging and analyzing the actions of the student on the software. These logs could also be used to improve the user interface. A third team in the research group is devoted to this task.

QED-Tutrix is a tool that has an immense potential for educational and research purposes. In its current state, it already provides interesting features that have generally been well received both by teachers and students. It is however crippled by the very limited selection of problems, with only five problems included currently. This limitation is due to the lengthiness of manually writing down all possible proofs for each problem. It is therefore a priority for us to handle that issue by automating the generation of proofs. The generated proofs must follow three specific criteria: readability, accessibility at a high-school level and adaptability to various deductive isles, in accordance to the didactical contract. We experimented briefly with the development of a very simple deduction engine, that offered promising results by successfully generating the inference graph for the two problems we encoded.

The next step is to confirm these results by generating the five inference graphs for the problems currently in QEDX, to ensure the validity of the method. Then, we must address the handling of inferential shortcuts by allowing the teacher to customize the allowed properties and the level of formality required from the students. Finally, we could explore possibilities to include the automated generation of intermediate constructions to the deductive engine.
 
\bibliographystyle{eptcs}
\bibliography{EPTCS-Biblio-QEDX}

\newcommand{\noopsort}[1]{} \newcommand{\singleletter}[1]{#1}
\begin{thebibliography}{10}
\providecommand{\bibitemdeclare}[2]{}
\providecommand{\surnamestart}{}
\providecommand{\surnameend}{}
\providecommand{\urlprefix}{Available at }
\providecommand{\url}[1]{\texttt{#1}}
\providecommand{\href}[2]{\texttt{#2}}
\providecommand{\urlalt}[2]{\href{#1}{#2}}
\providecommand{\doi}[1]{doi:\urlalt{http://dx.doi.org/#1}{#1}}
\providecommand{\bibinfo}[2]{#2}

\bibitemdeclare{misc}{geogebraAPI}
\bibitem{geogebraAPI}
\emph{\bibinfo{title}{GeoGebra API}}.
\newblock \urlprefix\url{http://dev.geogebra.org/trac/wiki/WikiStart}.

\bibitemdeclare{misc}{geometrix}
\bibitem{geometrix}
\emph{\bibinfo{title}{Géométrix}}.
\newblock \urlprefix\url{http://geometrix.free.fr/site/}.

\bibitemdeclare{misc}{mathway}
\bibitem{mathway}
\emph{\bibinfo{title}{Mathway {\textbar} {Math} {Problem} {Solver}}}.
\newblock \urlprefix\url{https://www.mathway.com/Algebra}.

\bibitemdeclare{inproceedings}{aleven1998combatting}
\bibitem{aleven1998combatting}
\bibinfo{author}{Vincent \surnamestart Aleven\surnameend},
  \bibinfo{author}{Kenneth \surnamestart Koedinger\surnameend},
  \bibinfo{author}{H~\surnamestart Colleen~Sinclair\surnameend} \&
  \bibinfo{author}{Jaclyn \surnamestart Snyder\surnameend}
  (\bibinfo{year}{1998}): \emph{\bibinfo{title}{Combatting shallow learning in
  a tutor for geometry problem solving}}.
\newblock In: {\sl \bibinfo{booktitle}{Intelligent Tutoring Systems}},
  \bibinfo{organization}{Springer}, pp. \bibinfo{pages}{364--373},
  \doi{10.1007/3-540-68716-5_42}.

\bibitemdeclare{inproceedings}{aleven2000limitations}
\bibitem{aleven2000limitations}
\bibinfo{author}{Vincent \surnamestart Aleven\surnameend} \&
  \bibinfo{author}{Kenneth~R \surnamestart Koedinger\surnameend}
  (\bibinfo{year}{2000}): \emph{\bibinfo{title}{Limitations of student control:
  Do students know when they need help?}}
\newblock In: {\sl \bibinfo{booktitle}{Intelligent tutoring systems}},
  \bibinfo{volume}{1839}, \bibinfo{organization}{Springer}, pp.
  \bibinfo{pages}{292--303}, \doi{10.1007/3-540-45108-0_33}.

\bibitemdeclare{article}{aleven2013knowledge}
\bibitem{aleven2013knowledge}
\bibinfo{author}{Vincent \surnamestart Aleven\surnameend} \&
  \bibinfo{author}{Kenneth~R \surnamestart Koedinger\surnameend}
  (\bibinfo{year}{2013}): \emph{\bibinfo{title}{Knowledge Component (KC)
  Approaches to Learner Modeling}}.
\newblock {\sl \bibinfo{journal}{Design Recommendations for Intelligent
  Tutoring Systems}} \bibinfo{volume}{1}, pp. \bibinfo{pages}{165--182}.

\bibitemdeclare{article}{aleven2006toward}
\bibitem{aleven2006toward}
\bibinfo{author}{Vincent \surnamestart Aleven\surnameend},
  \bibinfo{author}{Bruce \surnamestart Mclaren\surnameend},
  \bibinfo{author}{Ido \surnamestart Roll\surnameend} \&
  \bibinfo{author}{Kenneth \surnamestart Koedinger\surnameend}
  (\bibinfo{year}{2006}): \emph{\bibinfo{title}{Toward meta-cognitive tutoring:
  A model of help seeking with a Cognitive Tutor}}.
\newblock {\sl \bibinfo{journal}{International Journal of Artificial
  Intelligence in Education}} \bibinfo{volume}{16}(\bibinfo{number}{2}), pp.
  \bibinfo{pages}{101--128}.

\bibitemdeclare{article}{aleven2002effective}
\bibitem{aleven2002effective}
\bibinfo{author}{Vincent~AWMM \surnamestart Aleven\surnameend} \&
  \bibinfo{author}{Kenneth~R \surnamestart Koedinger\surnameend}
  (\bibinfo{year}{2002}): \emph{\bibinfo{title}{An effective metacognitive
  strategy: Learning by doing and explaining with a computer-based Cognitive
  Tutor}}.
\newblock {\sl \bibinfo{journal}{Cognitive science}}
  \bibinfo{volume}{26}(\bibinfo{number}{2}), pp. \bibinfo{pages}{147--179},
  \doi{10.1207/s15516709cog2602_1}.

\bibitemdeclare{article}{anderson1996act}
\bibitem{anderson1996act}
\bibinfo{author}{John~R \surnamestart Anderson\surnameend}
  (\bibinfo{year}{1996}): \emph{\bibinfo{title}{ACT: A simple theory of complex
  cognition.}}
\newblock {\sl \bibinfo{journal}{American Psychologist}}
  \bibinfo{volume}{51}(\bibinfo{number}{4}), p. \bibinfo{pages}{355},
  \doi{10.1037/0003-066X.51.4.355}.

\bibitemdeclare{article}{anderson2000implications}
\bibitem{anderson2000implications}
\bibinfo{author}{John~R \surnamestart Anderson\surnameend} \&
  \bibinfo{author}{C~\surnamestart Schunn\surnameend} (\bibinfo{year}{2000}):
  \emph{\bibinfo{title}{Implications of the ACT-R learning theory: No magic
  bullets}}.
\newblock {\sl \bibinfo{journal}{Advances in instructional psychology,
  Educational design and cognitive science}}, pp. \bibinfo{pages}{1--33}.

\bibitemdeclare{inproceedings}{arroyo2004web}
\bibitem{arroyo2004web}
\bibinfo{author}{Ivon \surnamestart Arroyo\surnameend}, \bibinfo{author}{Carole
  \surnamestart Beal\surnameend}, \bibinfo{author}{Tom \surnamestart
  Murray\surnameend}, \bibinfo{author}{Rena \surnamestart Walles\surnameend} \&
  \bibinfo{author}{Beverly~P \surnamestart Woolf\surnameend}
  (\bibinfo{year}{2004}): \emph{\bibinfo{title}{Web-based intelligent
  multimedia tutoring for high stakes achievement tests}}.
\newblock In: {\sl \bibinfo{booktitle}{Intelligent Tutoring Systems}},
  \bibinfo{organization}{Springer}, pp. \bibinfo{pages}{142--169},
  \doi{10.1007/978-3-540-30139-4_44}.

\bibitemdeclare{article}{balacheff2003baghera}
\bibitem{balacheff2003baghera}
\bibinfo{author}{Nicolas \surnamestart Balacheff\surnameend},
  \bibinfo{author}{Ricardo \surnamestart Caferra\surnameend},
  \bibinfo{author}{Michele \surnamestart Cerulli\surnameend},
  \bibinfo{author}{Nathalie \surnamestart Gaudin\surnameend},
  \bibinfo{author}{Mirko \surnamestart Maracci\surnameend},
  \bibinfo{author}{Maria~Alessandra \surnamestart Mariotti\surnameend},
  \bibinfo{author}{Jean-Pierre \surnamestart Muller\surnameend},
  \bibinfo{author}{Jean-Fran{\c{c}}ois \surnamestart Nicaud\surnameend},
  \bibinfo{author}{Michel \surnamestart Occello\surnameend},
  \bibinfo{author}{Federica \surnamestart Olivero\surnameend} et~al.
  (\bibinfo{year}{2003}): \emph{\bibinfo{title}{Baghera Assessment Project,
  designing an hybrid and emergent educational society}}.

\bibitemdeclare{phdthesis}{baulac1990micromonde}
\bibitem{baulac1990micromonde}
\bibinfo{author}{Yves \surnamestart Baulac\surnameend} (\bibinfo{year}{1990}):
  \emph{\bibinfo{title}{Un micromonde de g{\'e}om{\'e}trie,
  Cabri-g{\'e}om{\`e}tre}}.
\newblock Ph.D. thesis, \bibinfo{school}{Universit{\'e} Joseph-Fourier-Grenoble
  I}.

\bibitemdeclare{article}{botana2015automated}
\bibitem{botana2015automated}
\bibinfo{author}{Francisco \surnamestart Botana\surnameend},
  \bibinfo{author}{Markus \surnamestart Hohenwarter\surnameend},
  \bibinfo{author}{Predrag \surnamestart Jani{\v{c}}i{\'c}\surnameend},
  \bibinfo{author}{Zolt{\'a}n \surnamestart Kov{\'a}cs\surnameend},
  \bibinfo{author}{Ivan \surnamestart Petrovi{\'c}\surnameend},
  \bibinfo{author}{Tom{\'a}s \surnamestart Recio\surnameend} \&
  \bibinfo{author}{Simon \surnamestart Weitzhofer\surnameend}
  (\bibinfo{year}{2015}): \emph{\bibinfo{title}{Automated theorem proving in
  GeoGebra: Current achievements}}.
\newblock {\sl \bibinfo{journal}{Journal of Automated Reasoning}}
  \bibinfo{volume}{55}(\bibinfo{number}{1}), pp. \bibinfo{pages}{39--59},
  \doi{10.1007/s10817-015-9326-4}.

\bibitemdeclare{inproceedings}{boutry2016tarski}
\bibitem{boutry2016tarski}
\bibinfo{author}{Pierre \surnamestart Boutry\surnameend},
  \bibinfo{author}{Gabriel \surnamestart Braun\surnameend} \&
  \bibinfo{author}{Julien \surnamestart Narboux\surnameend}
  (\bibinfo{year}{2016}): \emph{\bibinfo{title}{From Tarski to Descartes:
  formalization of the arithmetization of euclidean geometry}}.
\newblock In: {\sl \bibinfo{booktitle}{SCSS 2016, the 7th International
  Symposium on Symbolic Computation in Software Science}},
  \bibinfo{volume}{39}, \bibinfo{organization}{EasyChair},
  p.~\bibinfo{pages}{15}, \doi{10.29007/k47p}.

\bibitemdeclare{article}{braun2017synthetic}
\bibitem{braun2017synthetic}
\bibinfo{author}{Gabriel \surnamestart Braun\surnameend} \&
  \bibinfo{author}{Julien \surnamestart Narboux\surnameend}
  (\bibinfo{year}{2017}): \emph{\bibinfo{title}{A synthetic proof of Pappus’
  theorem in Tarski’s geometry}}.
\newblock {\sl \bibinfo{journal}{Journal of Automated Reasoning}}
  \bibinfo{volume}{58}(\bibinfo{number}{2}), pp. \bibinfo{pages}{209--230},
  \doi{10.1007/s10817-016-9374-4}.

\bibitemdeclare{book}{brousseau2006theory}
\bibitem{brousseau2006theory}
\bibinfo{author}{Guy \surnamestart Brousseau\surnameend}
  (\bibinfo{year}{2006}): \emph{\bibinfo{title}{Theory of didactical situations
  in mathematics: Didactique des math{\'e}matiques, 1970--1990}}.
\newblock \bibinfo{volume}{19}, \bibinfo{publisher}{Springer Science \&
  Business Media}, \doi{10.1007/0-306-47211-2}.

\bibitemdeclare{incollection}{buchberger1988applications}
\bibitem{buchberger1988applications}
\bibinfo{author}{Bruno \surnamestart Buchberger\surnameend}
  (\bibinfo{year}{1988}): \emph{\bibinfo{title}{Applications of Gr{\"o}bner
  bases in non-linear computational geometry}}.
\newblock In: {\sl \bibinfo{booktitle}{Trends in computer algebra}},
  \bibinfo{publisher}{Springer}, pp. \bibinfo{pages}{52--80},
  \doi{10.1007/3-540-18928-9_5}.

\bibitemdeclare{inproceedings}{chou1993automated}
\bibitem{chou1993automated}
\bibinfo{author}{S-C \surnamestart Chou\surnameend}, \bibinfo{author}{X-S
  \surnamestart Gao\surnameend} \& \bibinfo{author}{J-Z \surnamestart
  Zhang\surnameend} (\bibinfo{year}{1993}): \emph{\bibinfo{title}{Automated
  production of traditional proofs for constructive geometry theorems}}.
\newblock In: {\sl \bibinfo{booktitle}{Logic in Computer Science, 1993.
  LICS'93., Proceedings of Eighth Annual IEEE Symposium on}},
  \bibinfo{organization}{IEEE}, pp. \bibinfo{pages}{48--56},
  \doi{10.1109/LICS.1993.287601}.

\bibitemdeclare{article}{chou1988introduction}
\bibitem{chou1988introduction}
\bibinfo{author}{Shang-Ching \surnamestart Chou\surnameend}
  (\bibinfo{year}{1988}): \emph{\bibinfo{title}{An introduction to Wu's method
  for mechanical theorem proving in geometry}}.
\newblock {\sl \bibinfo{journal}{Journal of Automated Reasoning}}
  \bibinfo{volume}{4}(\bibinfo{number}{3}), pp. \bibinfo{pages}{237--267},
  \doi{10.1007/BF00244942}.

\bibitemdeclare{book}{chou1994machine}
\bibitem{chou1994machine}
\bibinfo{author}{Shang-Ching \surnamestart Chou\surnameend},
  \bibinfo{author}{Xiao-Shan \surnamestart Gao\surnameend} \&
  \bibinfo{author}{Jing-Zhong \surnamestart Zhang\surnameend}
  (\bibinfo{year}{1994}): \emph{\bibinfo{title}{Machine proofs in geometry:
  Automated production of readable proofs for geometry theorems}}.
\newblock \bibinfo{volume}{6}, \bibinfo{publisher}{World Scientific},
  \doi{10.1142/9789812798152_0002}.

\bibitemdeclare{article}{chou1996automated}
\bibitem{chou1996automated}
\bibinfo{author}{Shang-Ching \surnamestart Chou\surnameend},
  \bibinfo{author}{Xiao-Shan \surnamestart Gao\surnameend} \&
  \bibinfo{author}{Jing-Zhong \surnamestart Zhang\surnameend}
  (\bibinfo{year}{1996}): \emph{\bibinfo{title}{Automated generation of
  readable proofs with geometric invariants. II. Theorem proving with
  full-angles}}.
\newblock {\sl \bibinfo{journal}{Journal of Automated Reasoning}}
  \bibinfo{volume}{17}(\bibinfo{number}{3}), pp. \bibinfo{pages}{349--370},
  \doi{10.1007/BF00283134}.

\bibitemdeclare{article}{coelho1986automated}
\bibitem{coelho1986automated}
\bibinfo{author}{Helder \surnamestart Coelho\surnameend} \&
  \bibinfo{author}{Luis~Moniz \surnamestart Pereira\surnameend}
  (\bibinfo{year}{1986}): \emph{\bibinfo{title}{Automated reasoning in geometry
  theorem proving with Prolog}}.
\newblock {\sl \bibinfo{journal}{Journal of Automated Reasoning}}
  \bibinfo{volume}{2}(\bibinfo{number}{4}), pp. \bibinfo{pages}{329--390},
  \doi{10.1007/BF00248249}.

\bibitemdeclare{inproceedings}{el2005development}
\bibitem{el2005development}
\bibinfo{author}{Simon \surnamestart El-Khoury\surnameend},
  \bibinfo{author}{Philippe~R \surnamestart Richard\surnameend},
  \bibinfo{author}{Esma \surnamestart A{\"\i}meur\surnameend} \&
  \bibinfo{author}{Josep~M \surnamestart Fortuny\surnameend}
  (\bibinfo{year}{2005}): \emph{\bibinfo{title}{Development of an Intelligent
  Tutorial System to Enhance Students’ Mathematical Competence in Problem
  Solving}}.
\newblock In: {\sl \bibinfo{booktitle}{E-Learn: World Conference on E-Learning
  in Corporate, Government, Healthcare, and Higher Education}},
  \bibinfo{organization}{Association for the Advancement of Computing in
  Education (AACE)}, pp. \bibinfo{pages}{2042--2049}.

\bibitemdeclare{article}{elcock1977representation}
\bibitem{elcock1977representation}
\bibinfo{author}{EW~\surnamestart Elcock\surnameend} (\bibinfo{year}{1977}):
  \emph{\bibinfo{title}{Representation of knowledge in geometry machine}}.
\newblock {\sl \bibinfo{journal}{Machine Intelligence}} \bibinfo{volume}{8},
  pp. \bibinfo{pages}{11--29}.

\bibitemdeclare{incollection}{falmagne2006assessment}
\bibitem{falmagne2006assessment}
\bibinfo{author}{Jean-Claude \surnamestart Falmagne\surnameend},
  \bibinfo{author}{Eric \surnamestart Cosyn\surnameend},
  \bibinfo{author}{Jean-Paul \surnamestart Doignon\surnameend} \&
  \bibinfo{author}{Nicolas \surnamestart Thi{\'e}ry\surnameend}
  (\bibinfo{year}{2006}): \emph{\bibinfo{title}{The assessment of knowledge, in
  theory and in practice}}.
\newblock In: {\sl \bibinfo{booktitle}{Formal concept analysis}},
  \bibinfo{publisher}{Springer}, pp. \bibinfo{pages}{61--79},
  \doi{10.1007/11671404_4}.

\bibitemdeclare{inproceedings}{gelernter1960empirical}
\bibitem{gelernter1960empirical}
\bibinfo{author}{Herbert \surnamestart Gelernter\surnameend},
  \bibinfo{author}{James~R \surnamestart Hansen\surnameend} \&
  \bibinfo{author}{Donald~W \surnamestart Loveland\surnameend}
  (\bibinfo{year}{1960}): \emph{\bibinfo{title}{Empirical explorations of the
  geometry theorem machine}}.
\newblock In: {\sl \bibinfo{booktitle}{Papers presented at the May 3-5, 1960,
  western joint IRE-AIEE-ACM computer conference}},
  \bibinfo{organization}{ACM}, pp. \bibinfo{pages}{143--149}.

\bibitemdeclare{article}{greeno1979constructions}
\bibitem{greeno1979constructions}
\bibinfo{author}{James~G \surnamestart Greeno\surnameend}
  (\bibinfo{year}{1979}): \emph{\bibinfo{title}{Constructions in Geometry
  Problem Solving.}}
\newblock \doi{10.3758/BF03198261}.

\bibitemdeclare{article}{hohenwarter2013geogebra}
\bibitem{hohenwarter2013geogebra}
\bibinfo{author}{Markus \surnamestart Hohenwarter\surnameend}
  (\bibinfo{year}{2013}): \emph{\bibinfo{title}{GeoGebra 4.4--From desktops to
  tablets}}.
\newblock {\sl \bibinfo{journal}{Indagatio Didactica}}
  \bibinfo{volume}{5}(\bibinfo{number}{1}).

\bibitemdeclare{inproceedings}{janicic2006gclc}
\bibitem{janicic2006gclc}
\bibinfo{author}{Predrag \surnamestart Janicic\surnameend}
  (\bibinfo{year}{2006}): \emph{\bibinfo{title}{GCLC-A Tool for constructive
  euclidean geometry and more than that}}.
\newblock In: {\sl \bibinfo{booktitle}{ICMS}},
  \bibinfo{organization}{Springer}, pp. \bibinfo{pages}{58--73},
  \doi{10.1007/11832225_6}.

\bibitemdeclare{phdthesis}{jean2000pepite}
\bibitem{jean2000pepite}
\bibinfo{author}{St{\'e}phanie \surnamestart Jean-Daubias\surnameend}
  (\bibinfo{year}{2000}): \emph{\bibinfo{title}{P{\'E}PITE: un syst{\`e}me
  d'assistance au diagnostic de comp{\'e}tences}}.
\newblock Ph.D. thesis, \bibinfo{school}{Universit{\'e} du Maine}.

\bibitemdeclare{article}{kapur1986using}
\bibitem{kapur1986using}
\bibinfo{author}{Deepak \surnamestart Kapur\surnameend} (\bibinfo{year}{1986}):
  \emph{\bibinfo{title}{Using Gr{\"o}bner bases to reason about geometry
  problems}}.
\newblock {\sl \bibinfo{journal}{Journal of Symbolic Computation}}
  \bibinfo{volume}{2}(\bibinfo{number}{4}), pp. \bibinfo{pages}{399--408},
  \doi{10.1016/S0747-7171(86)80007-4}.

\bibitemdeclare{inproceedings}{koedinger1991design}
\bibitem{koedinger1991design}
\bibinfo{author}{K~\surnamestart Koedinger\surnameend} (\bibinfo{year}{1991}):
  \emph{\bibinfo{title}{On the design of novel notations and actions to
  facilitate thinking and learning}}.
\newblock In: {\sl \bibinfo{booktitle}{Proceedings of the International
  Conference on the Learning Sciences}}, pp. \bibinfo{pages}{266--273}.

\bibitemdeclare{article}{koedinger1990abstract}
\bibitem{koedinger1990abstract}
\bibinfo{author}{Kenneth~R \surnamestart Koedinger\surnameend} \&
  \bibinfo{author}{John~R \surnamestart Anderson\surnameend}
  (\bibinfo{year}{1990}): \emph{\bibinfo{title}{Abstract planning and
  perceptual chunks: Elements of expertise in geometry}}.
\newblock {\sl \bibinfo{journal}{Cognitive Science}}
  \bibinfo{volume}{14}(\bibinfo{number}{4}), pp. \bibinfo{pages}{511--550},
  \doi{10.1207/s15516709cog1404_2}.

\bibitemdeclare{article}{koedinger1993effective}
\bibitem{koedinger1993effective}
\bibinfo{author}{Kenneth~R \surnamestart Koedinger\surnameend} \&
  \bibinfo{author}{John~R \surnamestart Anderson\surnameend}
  (\bibinfo{year}{1993}): \emph{\bibinfo{title}{Effective use of intelligent
  software in high school math classrooms}}.

\bibitemdeclare{inproceedings}{kordaki2006potential}
\bibitem{kordaki2006potential}
\bibinfo{author}{Maria \surnamestart Kordaki\surnameend} \&
  \bibinfo{author}{Alexios \surnamestart Mastrogiannis\surnameend}
  (\bibinfo{year}{2006}): \emph{\bibinfo{title}{The potential of
  multiple-solution tasks in e-learning environments: Exploiting the tools of
  Cabri Geometry II}}.
\newblock In: {\sl \bibinfo{booktitle}{E-Learn: World Conference on E-Learning
  in Corporate, Government, Healthcare, and Higher Education}},
  \bibinfo{organization}{Association for the Advancement of Computing in
  Education (AACE)}, pp. \bibinfo{pages}{97--104}.

\bibitemdeclare{article}{kuzniak2014espacios}
\bibitem{kuzniak2014espacios}
\bibinfo{author}{Alain \surnamestart Kuzniak\surnameend} \&
  \bibinfo{author}{Philippe~R \surnamestart Richard\surnameend}
  (\bibinfo{year}{2014}): \emph{\bibinfo{title}{Espacios de trabajo
  matem{\'a}tico. Puntos de vista y perspectivas}}.
\newblock {\sl \bibinfo{journal}{Revista latinoamericana de investigaci{\'o}n
  en matem{\'a}tica educativa}} \bibinfo{volume}{17}(\bibinfo{number}{4}),
  \doi{10.12802/relime.13.1741a}.

\bibitemdeclare{phdthesis}{Leduc2016}
\bibitem{Leduc2016}
\bibinfo{author}{N.~\surnamestart Leduc\surnameend} (\bibinfo{year}{2016}):
  \emph{\bibinfo{title}{QED-Tutrix : système tutoriel intelligent pour
  l’accompagnement d’élèves en situation de résolution de problèmes de
  démonstration en géométrie plane}}.
\newblock Ph.D. thesis, \bibinfo{school}{École polytechnique de Montréal.}

\bibitemdeclare{article}{luengo1997cabri}
\bibitem{luengo1997cabri}
\bibinfo{author}{Vanda \surnamestart Luengo\surnameend} (\bibinfo{year}{1997}):
  \emph{\bibinfo{title}{Cabri-Euclide: Un micromonde de Preuve int{\'e}grant la
  r{\'e}futation}}.
\newblock {\sl \bibinfo{journal}{These de doctorat, INPG, France}}.

\bibitemdeclare{article}{luengo2005some}
\bibitem{luengo2005some}
\bibinfo{author}{Vanda \surnamestart Luengo\surnameend} (\bibinfo{year}{2005}):
  \emph{\bibinfo{title}{Some didactical and epistemological considerations in
  the design of educational software: the Cabri-Euclide example}}.
\newblock {\sl \bibinfo{journal}{International Journal of Computers for
  Mathematical Learning}} \bibinfo{volume}{10}(\bibinfo{number}{1}), pp.
  \bibinfo{pages}{1--29}, \doi{10.1007/s10758-005-4580-x}.

\bibitemdeclare{article}{luengo1998contraintes}
\bibitem{luengo1998contraintes}
\bibinfo{author}{Vanda \surnamestart Luengo\surnameend} \&
  \bibinfo{author}{Nicolas \surnamestart Balacheff\surnameend}
  (\bibinfo{year}{1998}): \emph{\bibinfo{title}{Contraintes informatiques et
  environnements d'apprentissage de la d{\'e}monstration en
  math{\'e}matiques.}}
\newblock {\sl \bibinfo{journal}{Sciences et Techniques Educatives}}
  \bibinfo{volume}{5}, pp. \bibinfo{pages}{15--45}.

\bibitemdeclare{article}{matsuda2004gramy}
\bibitem{matsuda2004gramy}
\bibinfo{author}{Noboru \surnamestart Matsuda\surnameend} \&
  \bibinfo{author}{Kurt \surnamestart Vanlehn\surnameend}
  (\bibinfo{year}{2004}): \emph{\bibinfo{title}{Gramy: A geometry theorem
  prover capable of construction}}.
\newblock {\sl \bibinfo{journal}{Journal of Automated Reasoning}}
  \bibinfo{volume}{32}(\bibinfo{number}{1}), pp. \bibinfo{pages}{3--33},
  \doi{10.1023/B:JARS.0000021960.39761.b7}.

\bibitemdeclare{inproceedings}{matsuda2005advanced}
\bibitem{matsuda2005advanced}
\bibinfo{author}{Noboru \surnamestart Matsuda\surnameend} \&
  \bibinfo{author}{Kurt \surnamestart VanLehn\surnameend}
  (\bibinfo{year}{2005}): \emph{\bibinfo{title}{Advanced Geometry Tutor: An
  intelligent tutor that teaches proof-writing with construction.}}
\newblock In: {\sl \bibinfo{booktitle}{AIED}}, \bibinfo{volume}{125}, pp.
  \bibinfo{pages}{443--450}.

\bibitemdeclare{article}{melis2009activemath}
\bibitem{melis2009activemath}
\bibinfo{author}{Erica \surnamestart Melis\surnameend}, \bibinfo{author}{Giorgi
  \surnamestart Goguadze\surnameend}, \bibinfo{author}{Paul \surnamestart
  Libbrecht\surnameend} \& \bibinfo{author}{Carsten \surnamestart
  Ullrich\surnameend} (\bibinfo{year}{2009}):
  \emph{\bibinfo{title}{Activemath--a learning platform with semantic web
  features}}.
\newblock {\sl \bibinfo{journal}{The Future of Learning}}, p.
  \bibinfo{pages}{159}.

\bibitemdeclare{inproceedings}{narboux2006mechanical}
\bibitem{narboux2006mechanical}
\bibinfo{author}{Julien \surnamestart Narboux\surnameend}
  (\bibinfo{year}{2006}): \emph{\bibinfo{title}{Mechanical theorem proving in
  Tarski’s geometry}}.
\newblock In: {\sl \bibinfo{booktitle}{International Workshop on Automated
  Deduction in Geometry}}, \bibinfo{organization}{Springer}, pp.
  \bibinfo{pages}{139--156}, \doi{10.1007/978-3-540-77356-6_9}.

\bibitemdeclare{article}{nevins1975plane}
\bibitem{nevins1975plane}
\bibinfo{author}{Arthur~J \surnamestart Nevins\surnameend}
  (\bibinfo{year}{1975}): \emph{\bibinfo{title}{Plane geometry theorem proving
  using forward chaining}}.
\newblock {\sl \bibinfo{journal}{Artificial Intelligence}}
  \bibinfo{volume}{6}(\bibinfo{number}{1}), pp. \bibinfo{pages}{1--23},
  \doi{10.1016/0004-3702(75)90013-2}.

\bibitemdeclare{article}{py1994reconnaissance}
\bibitem{py1994reconnaissance}
\bibinfo{author}{D~\surnamestart Py\surnameend} (\bibinfo{year}{1994}):
  \emph{\bibinfo{title}{Reconnaissance de plan pour la mod{\'e}lisation de
  l'{\'e}l{\`e}ve. Le projet Mentoniezh}}.
\newblock {\sl \bibinfo{journal}{Recherches en didactique des
  math{\'e}matiques}} \bibinfo{volume}{14}(\bibinfo{number}{1/2}), pp.
  \bibinfo{pages}{113--138}.

\bibitemdeclare{article}{py1996aide}
\bibitem{py1996aide}
\bibinfo{author}{Dominique \surnamestart Py\surnameend} (\bibinfo{year}{1996}):
  \emph{\bibinfo{title}{Aide {\`a} la d{\'e}monstration en g{\'e}om{\'e}trie:
  le projet Mentoniezh}}.
\newblock {\sl \bibinfo{journal}{Sciences et techniques {\'e}ducatives}}
  \bibinfo{volume}{3}(\bibinfo{number}{2}), pp. \bibinfo{pages}{227--256}.

\bibitemdeclare{article}{py2001environnements}
\bibitem{py2001environnements}
\bibinfo{author}{Dominique \surnamestart Py\surnameend} (\bibinfo{year}{2001}):
  \emph{\bibinfo{title}{Environnements interactifs d'apprentissage et
  d{\'e}monstration en g{\'e}om{\'e}trie}}.

\bibitemdeclare{book}{rabardel1995hommes}
\bibitem{rabardel1995hommes}
\bibinfo{author}{Pierre \surnamestart Rabardel\surnameend}
  (\bibinfo{year}{1995}): \emph{\bibinfo{title}{Les hommes et les technologies;
  approche cognitive des instruments contemporains}}.
\newblock \bibinfo{publisher}{Armand Colin}.

\bibitemdeclare{article}{richard2004inference}
\bibitem{richard2004inference}
\bibinfo{author}{Philippe~R \surnamestart Richard\surnameend}
  (\bibinfo{year}{2004}): \emph{\bibinfo{title}{L'inf{\'e}rence figurale: un
  pas de raisonnement discursivo-graphique}}.
\newblock {\sl \bibinfo{journal}{Educational Studies in Mathematics}}
  \bibinfo{volume}{57}(\bibinfo{number}{2}), pp. \bibinfo{pages}{229--263},
  \doi{10.1023/B:EDUC.0000049272.75852.c4}.

\bibitemdeclare{inproceedings}{richard2007amelioration}
\bibitem{richard2007amelioration}
\bibinfo{author}{Philippe~R \surnamestart Richard\surnameend} \&
  \bibinfo{author}{Josep~M \surnamestart Fortuny\surnameend}
  (\bibinfo{year}{2007}): \emph{\bibinfo{title}{Am{\'e}lioration des
  comp{\'e}tences argumentatives {\`a} l’aide d’un syst{\`e}me tutoriel en
  classe de math{\'e}matique au secondaire}}.
\newblock In: {\sl \bibinfo{booktitle}{Annales de didactique et de sciences
  cognitives}}, \bibinfo{volume}{12}, pp. \bibinfo{pages}{83--116}.

\bibitemdeclare{article}{richard2011didactic}
\bibitem{richard2011didactic}
\bibinfo{author}{Philippe~R \surnamestart Richard\surnameend},
  \bibinfo{author}{Josep~Maria \surnamestart Fortuny\surnameend},
  \bibinfo{author}{Michel \surnamestart Gagnon\surnameend},
  \bibinfo{author}{Nicolas \surnamestart Leduc\surnameend},
  \bibinfo{author}{Eloi \surnamestart Puertas\surnameend} \&
  \bibinfo{author}{Mich{\`e}le \surnamestart Tessier-Baillargeon\surnameend}
  (\bibinfo{year}{2011}): \emph{\bibinfo{title}{Didactic and theoretical-based
  perspectives in the experimental development of an intelligent tutorial
  system for the learning of geometry}}.
\newblock {\sl \bibinfo{journal}{ZDM}}
  \bibinfo{volume}{43}(\bibinfo{number}{3}), pp. \bibinfo{pages}{425--439},
  \doi{10.1007/s11858-011-0320-y}.

\bibitemdeclare{inproceedings}{richard2017connectedness}
\bibitem{richard2017connectedness}
\bibinfo{author}{P.R. \surnamestart Richard\surnameend},
  \bibinfo{author}{Michel \surnamestart Gagnon\surnameend} \&
  \bibinfo{author}{J.~M. \surnamestart Fortuny\surnameend}
  (\bibinfo{year}{2018}): \emph{\bibinfo{title}{Connectedness of Problems and
  Impass Resolution in the Solving Process in Geometry: a Major Educational
  Challenge}}.
\newblock In \bibinfo{editor}{P.~\surnamestart Herbst\surnameend},
  \bibinfo{editor}{U.H. \surnamestart Cheah\surnameend},
  \bibinfo{editor}{K.~\surnamestart Jones\surnameend} \& \bibinfo{editor}{P.R.
  \surnamestart Richard\surnameend}, editors: {\sl
  \bibinfo{booktitle}{International Perspectives on the Teaching and Learning
  of Geometry in Secondary Schools}}, \bibinfo{organization}{Springer}, pp.
  \bibinfo{pages}{311--327}.

\bibitemdeclare{article}{roll2014benefits}
\bibitem{roll2014benefits}
\bibinfo{author}{Ido \surnamestart Roll\surnameend}, \bibinfo{author}{Ryan SJ~d
  \surnamestart Baker\surnameend}, \bibinfo{author}{Vincent \surnamestart
  Aleven\surnameend} \& \bibinfo{author}{Kenneth~R \surnamestart
  Koedinger\surnameend} (\bibinfo{year}{2014}): \emph{\bibinfo{title}{On the
  benefits of seeking (and avoiding) help in online problem-solving
  environments}}.
\newblock {\sl \bibinfo{journal}{Journal of the Learning Sciences}}
  \bibinfo{volume}{23}(\bibinfo{number}{4}), pp. \bibinfo{pages}{537--560},
  \doi{10.1016/S0360-1315(99)00030-5}.

\bibitemdeclare{phdthesis}{Tessier-Baillargeon2015}
\bibitem{Tessier-Baillargeon2015}
\bibinfo{author}{M~\surnamestart Tessier-Baillargeon\surnameend}
  (\bibinfo{year}{2015}): \emph{\bibinfo{title}{GeoGebraTUTOR : Développement
  d’un système tutoriel autonome pour l’accompagnement d’élèves en
  situation de résolution de problèmes de démonstration en géométrie plane
  et genèse d’un espace de travail géométrique idoine}}.
\newblock Ph.D. thesis, \bibinfo{school}{Université de Montréal.}

\bibitemdeclare{inproceedings}{webber2001baghera}
\bibitem{webber2001baghera}
\bibinfo{author}{Carine \surnamestart Webber\surnameend},
  \bibinfo{author}{Loris \surnamestart Bergia\surnameend},
  \bibinfo{author}{Sylvie \surnamestart Pesty\surnameend} \&
  \bibinfo{author}{Nicolas \surnamestart Balacheff\surnameend}
  (\bibinfo{year}{2001}): \emph{\bibinfo{title}{The Baghera project: a
  multi-agent architecture for human learning}}.
\newblock In: {\sl \bibinfo{booktitle}{Workshop-Multi-Agent Architectures for
  Distributed Learning Environments. In Proceedings International Conference on
  AI and Education. San Antonio, Texas}}.

\bibitemdeclare{article}{weber2001elm}
\bibitem{weber2001elm}
\bibinfo{author}{Gerhard \surnamestart Weber\surnameend} \&
  \bibinfo{author}{Peter \surnamestart Brusilovsky\surnameend}
  (\bibinfo{year}{2001}): \emph{\bibinfo{title}{ELM-ART: An adaptive versatile
  system for Web-based instruction}}.
\newblock {\sl \bibinfo{journal}{International Journal of Artificial
  Intelligence in Education (IJAIED)}} \bibinfo{volume}{12}, pp.
  \bibinfo{pages}{351--384}.

\bibitemdeclare{article}{wu1979elementary}
\bibitem{wu1979elementary}
\bibinfo{author}{H~\surnamestart Wu\surnameend} (\bibinfo{year}{1979}):
  \emph{\bibinfo{title}{An elementary method in the study of nonnegative
  curvature}}.
\newblock {\sl \bibinfo{journal}{Acta Mathematica}}
  \bibinfo{volume}{142}(\bibinfo{number}{1}), pp. \bibinfo{pages}{57--78},
  \doi{10.1007/BF02395057}.

\bibitemdeclare{article}{zhang1990parallel}
\bibitem{zhang1990parallel}
\bibinfo{author}{Jingzhong \surnamestart Zhang\surnameend},
  \bibinfo{author}{Lu~\surnamestart Yang\surnameend} \& \bibinfo{author}{Mike
  \surnamestart Deng\surnameend} (\bibinfo{year}{1990}):
  \emph{\bibinfo{title}{The parallel numerical method of mechanical theorem
  proving}}.
\newblock {\sl \bibinfo{journal}{Theoretical Computer Science}}
  \bibinfo{volume}{74}(\bibinfo{number}{3}), pp. \bibinfo{pages}{253--271},
  \doi{10.1016/0304-3975(90)90077-U}.

\end{thebibliography}

\end{document}